\documentclass{bmvc2k}

\usepackage{short}
\usepackage{algorithm}
\usepackage{algorithmic}
\usepackage{booktabs}
\usepackage{subfigure}

\newcommand{\method}{SOKD\xspace}
\newcommand{\kd}{knowledge distillation\xspace}

\title{Semi-Online Knowledge Distillation}

\addauthor{Zhiqiang Liu}{sezhiqiangliu@mail.scut.edu.cn}{1}
\addauthor{Yanxia Liu $^{*,} $}{cslyx@scut.edu.cn}{1}
\addauthor{Chengkai Huang}{fjswhck@hrbeu.edu.cn}{2}

\addinstitution{
School of Software\\
South China University of Technology\\
Guangzhou, China
}
\addinstitution{
College of Computer Science and Technology\\
Harbin Institute of Technology\\
Shenzhen, China
}

\runninghead{Liu \etal.}{Semi-Online Knowledge Distillation}

\def\eg{\emph{e.g}\bmvaOneDot}

\def\etal{\emph{et al}\bmvaOneDot}

\begin{document}

\maketitle

\begin{abstract}
Knowledge distillation is an effective and stable method for model compression via knowledge transfer. Conventional knowledge distillation (KD) is to transfer knowledge from a large and well pre-trained teacher network to a small student network, which is a one-way process. Recently, deep mutual learning (DML) has been proposed to help student networks learn collaboratively and simultaneously. However, to the best of our knowledge, KD and DML have never been jointly explored in a unified framework to solve the knowledge distillation problem. In this paper, we investigate that the teacher model supports more trustworthy supervision signals in KD, while the student captures more similar behaviors from the teacher in DML. Based on these observations, we first propose to combine KD with DML in a unified framework. Furthermore, we propose a Semi-Online Knowledge Distillation (\method) method that effectively improves the performance of the student and the teacher. In this method, we introduce the peer-teaching training fashion in DML in order to alleviate the student's imitation difficulty, and also leverage the supervision signals provided by the well-trained teacher in KD. Besides, we also show our framework can be easily extended to feature-based distillation methods. Extensive experiments on CIFAR-100 and ImageNet datasets demonstrate the proposed method achieves state-of-the-art performance. 
\end{abstract}

\section{Introduction}
In recent years, deep learning has demonstrated great success in many computer vision tasks such as image classification~\cite{he2016deep, huang2017densely, sandler2018mobilenetv2}, object detection~\cite{lin2017focal, ren2015faster, DBLP:conf/mm/XuZWTG20} and image generation~\cite{Goodfellow2014GenerativeAN, Liu2021CTSF, Zhuang2021ANS}, which mainly benefits from deeper and larger convolutional neural network architectures. However, these architectures require huge computation and massive storage, which makes them hard to deploy on resource-limited devices such as mobile phones and drones. Therefore, many recent studies have focused on network compression methods, including network pruning~\cite{he2017channel, zhuang2018discrimination, chin2020towards}, network quantization~\cite{wu2016quantized, Xie2020DeepTQ, wang2019haq}, and knowledge distillation~\cite{hinton2015distilling, romero2014fitnets, yuan2020revisiting}.

Knowledge distillation is a simple yet effective technology, which is orthogonal to other model compression methods. Existing distillation methods can be roughly divided into offline \kd and online \kd~\cite{DBLP:conf/cvpr/GuoWWYLHL20}. Without loss of generality, in this paper we investigate the typical offline and online \kd methods, namely KD~\cite{hinton2015distilling} and deep mutual learning (DML)~\cite{Zhang2018DeepML}. The key idea of KD is to improve the performance of a compact network (student) via the knowledge transfer from a large and well-trained network (teacher). Specifically, Hinton \etal~\cite{hinton2015distilling} regard the softened output probabilities of the teacher as knowledge and propose that the student absorbs this knowledge by approximating the teacher's outputs. Whereas, the student is generally hard to fully absorb the knowledge provided by the teacher in practice (\ie, fully match the outputs of the teacher). The main reason is that there exists a large performance gap between the converged large network and untrained small network~\cite{jin2019knowledge}. On the contrary, the networks in DML are easier to approach and learn from each other because they are regarded as students and trained simultaneously. Such peer-teaching training fashion makes the networks easily imitate the behaviors (\ie, representations and outputs) of each other. However, since all networks in DML are trained from scratch and their optimization directions seem unstable in the early training phase, there exist many unreliable supervision signals, which might lead to conflicts of networks' optimization~\cite{DBLP:conf/cvpr/GuoWWYLHL20}. Although both KD and DML exist some shortcomings, we argue that they are complementary to each other for knowledge distillation problem.

To verify the above viewpoint, we evaluate the imitation ability of the student~\footnote{For convenience, we only consider two networks case and refer to the smaller network in DML as student, otherwise as teacher.} based on CKA similarity~\cite{kornblith2019similarity} and two well-defined metrics which are imitation error rate (IER) and misleading rate (MR). We empirically found that 1) The peer-teaching training fashion in DML is a more effective way than one-way flow in KD for the student to learn similar behaviors of the teacher. 2) The well-trained teacher in KD can provide more trustworthy and stable supervision signals for the student than the teacher in DML. (See discussion in detail in section~\ref{sec:imitation}). 

Based on the above observations, in this paper we first propose a unified framework combining KD with DML, which effectively leverages the teacher's supervision signals in KD and alleviates the imitation difficulty via the peer-teaching training fashion in DML. To this end, we further propose a Semi-Online Knowledge Distillation (\method) method. In this method, we build a simple yet effective knowledge bridge module (KBM) to exploit the knowledge from the offline teacher and transfer it to the student online. Specifically, the KBM is the same as the high-level layers of the teacher while taking the output of the low-level layers as input. In the training phase, inspired by DML, we update the KBM and the student simultaneously while fixing the teacher. In the inference phase, we replace high-level layers with KBM to construct a new teacher. As a result, we obtain a compact but effective student as well as, surprisingly, a more powerful teacher.

Our contributions are summarized as follows:
\begin{itemize}
    \item We discover that KD and DML are complementary to each other on knowledge distillation task. Based on this observation, we propose a novel distillation method combining KD with DML. To the best of our knowledge, we are the first to jointly integrate KD with DML in a unified framework.
    \item We propose a Semi-Online Knowledge Distillation (\method) method that significantly improves the performance of the student and the teacher. To leverage the well-trained teacher's supervision signals and alleviate the imitation difficulty of the student, we propose a simple yet effective knowledge bridge module (KBM).
    \item We further extend our proposed framework to feature-based distillation methods. Extensive experiments on CIFAR-100 and ImageNet datasets demonstrate that the proposed \method not only outperforms state-of-the-art model distillation methods but also can be easily extended to feature-based distillation methods for improved learning accuracy.
\end{itemize}

\section{Related Work}

\textbf{Offline Knowledge Distillation.}
The training process of traditional offline knowledge distillation can be split into two stages: 1) pre-training a teacher; and 2) distilling the knowledge of teacher into student. Most existing work 
mainly concentrate on how to define and transfer knowledge. For example, Hinton \etal~\cite{hinton2015distilling} first propose to mimic softened logits of teacher.
Romero \etal~\cite{romero2014fitnets} propose to match the intermediate representations of the teacher and the student. After that, many studies investigate how to clearly define the knowledge based on the intermediate feature maps~\cite{Zagoruyko2017PayingMA, tung2019similarity}.
However, most feature-based distillation methods are sensitive to the structures of the teacher and the student~\cite{tian2019contrastive, xu2020knowledge}. Instead, in this paper we focus on improving the student's capacity to imitate the logits of the teacher. Such logit-based \kd can be well generalized to different teacher-student combinations.\\
\noindent \textbf{Online Knowledge Distillation.}
Zhang \etal~\cite{Zhang2018DeepML} propose an online distillation scheme called deep mutual learning (DML), in which all networks are treated as students and updated simultaneously. To reduce the computational cost, Lan \etal~\cite{Lan2018KnowledgeDB} propose a multi-branch network.
Recently, Chen \etal~\cite{chen2020online} introduce the self-attention mechanism to improve the diversity of students. However, the different outputs of students will conflict with each other, which may harm the convergence of training~\cite{DBLP:conf/cvpr/GuoWWYLHL20}. To improve the ability of converging with higher generalization, Guo \etal~\cite{DBLP:conf/cvpr/GuoWWYLHL20} generate a high-quality soft target via ensembling the output of students. Besides, some studies~\cite{zhang2019your, yun2020regularizing, hou2019learning} propose self-distillation methods, the special cases of online knowledge distillation, in which the teachers and students are the same networks. In this paper, we introduce the peer-teaching training fashion in DML into KD to simplify the difficulty for student to imitate teacher's behaviors.

\section{Proposed Method}
\subsection{Preliminary}
\textbf{Knowledge Distillation.}
The key idea of KD is that the student uses the softened output of the teacher as supervised signal for network training. Given the $i^{th}$ sample, we can get the logits $z_i=(z_i^1,z_i^2,...,z_i^C)$ before softmax function. Then the softened output probability of the teacher network is computed as:
\begin{equation}\label{eq:soft_target}
    p_i^t=\frac{exp(z_i^j/\tau)}{\sum_{j=1}^{C}exp(z_i^j/\tau)}, 
\end{equation}
where $\tau$ is a temperature factor. Similarly, the student produces a softened target $p_i^s$. The student mimics outputs of the teacher by minimizing Kullback-Leibler (KL) Divergence $KL(p^s, p^t)$. The total loss for the student network consists of the typical cross entropy loss $\mL_{ce}^s$ and the KL Divergence:
\begin{equation}\label{eq:kl}
    KL(p^s,p^t)=\sum_{x\sim \mD}p^s\log\frac{p^s}{p^t},  \mL^s=\lambda_1\mL_{ce}^s+\lambda_2KL(p^s,p^t),
\end{equation}
$\lambda_1$ and $\lambda_2$ are hyperparameters.

\textbf{Deep Mutual Learning.}
Zhang \etal~\cite{Zhang2018DeepML} propose an online knowledge distillation method named deep mutual learning, in which the teacher and the student networks are updated simultaneously. Specifically, the teacher and the student optimize their networks with their respective loss functions $\mL^t$ and $\mL^s$. The formulations are as follows:
\begin{equation}
\begin{split}
    \mL^t=\lambda_1\mL_{ce}^t+\lambda_2KL(p^t,p^s),  
    \mL^s=\lambda_1\mL_{ce}^s+\lambda_2KL(p^s,p^t), \\
\end{split}
\end{equation}
$\lambda_1$ and $\lambda_2$ are hyperparameters.

\subsection{Analysis on Imitation Ability of Student}
\label{sec:imitation}

In distillation methods, the student is expected to imitate the behaviors of the teacher, which means that the performance of the student largely depends on its imitation. To investigate the mimic ability of the student in KD and DML, we conduct experiments with the teacher-student pair of WRN-28-4 and WRN-16-2 on CIFAR-100 dataset. Specifically, we use CKA~\cite{kornblith2019similarity} to measure the representational similarity between the teacher and the student. The larger the CKA, the more similar the representations between teacher and student. Besides, we define the imitation error rate (IER) and the misleading rate (MR) as the metrics to measure the ability to imitate the final output. IER refers to the rate of different outputs between the teacher and the student. MR represents the rate of wrong outputs learned from the teacher. Let $\mD_{st}$ be the samples with the same prediction results between the teacher and the student in the training data $\mD$, $\mD_{sg}$ be the samples with the same prediction results between the ground-truth and the student in $\mD_{st}$. IER and MR are computed as $(1-|\mD_{st}|/|\mD|)\times100$ and $(1-|\mD_{sg}|/|\mD_{st}|)\times100$, respectively. $|\cdot|$ denotes number of samples.

From table~\ref{tab:similarity}, DML achieves much higher CKA and lower IER than KD, which means that the student in DML has stronger abilities to capture the representations of the teacher and mimic the output probabilities of the teacher effectively. However, DML achieves highest MR, meaning that the student in DML has learned much more wrong information from the teacher. The main reason is that in the early training phase the optimization direction of the teacher is unstable. The supervision signals provided by the teacher might be inaccurate and changing. Fortunately, we are able to easily obtain reliable supervision signals from the well-trained teacher in KD. Therefore, it is possible for the student to learn similar and accurate behaviors from the teacher and effectively improve its performance.
\begin{figure*}[t]
    \centering
	\includegraphics[width=0.85\linewidth]{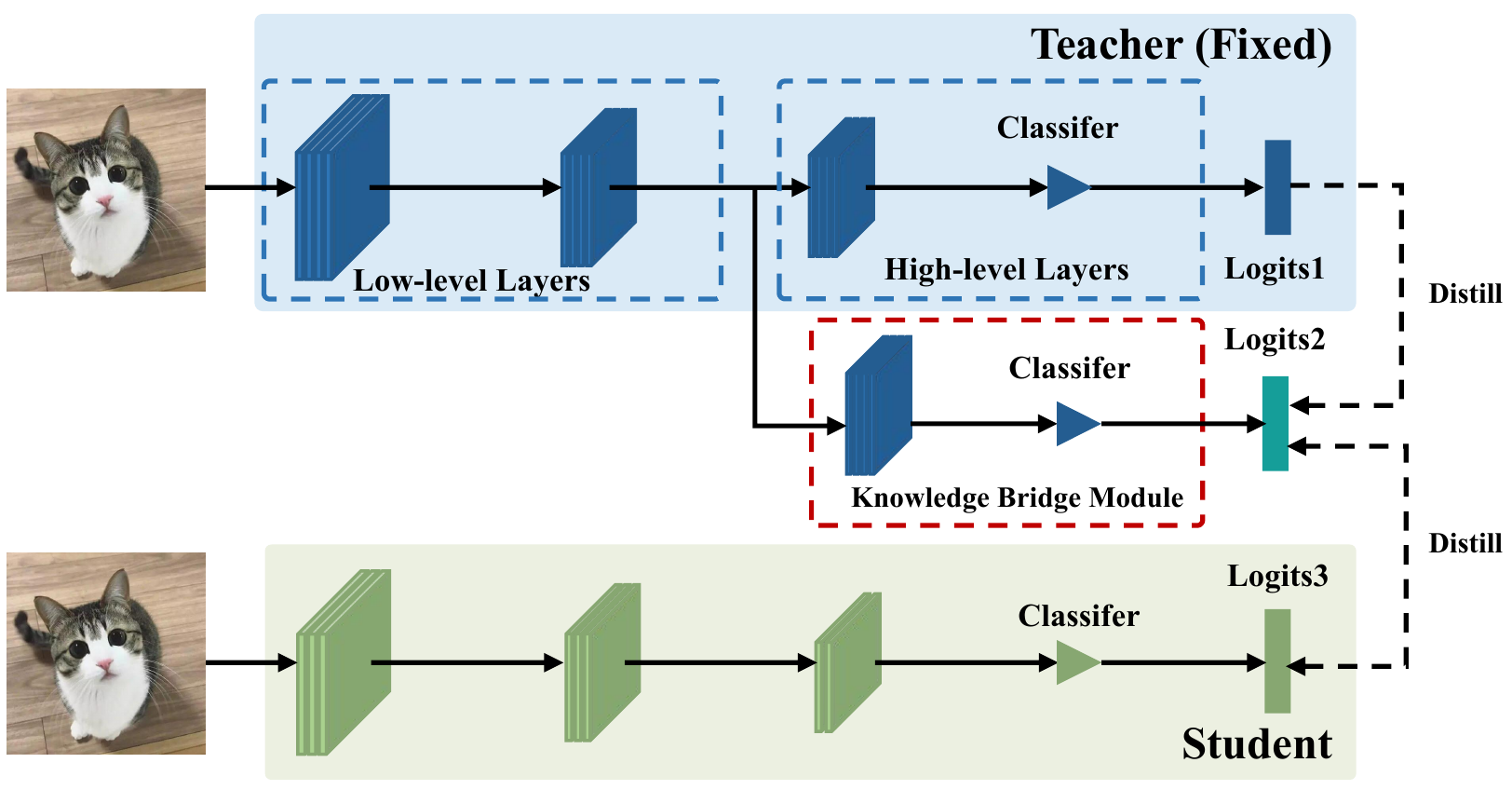}
	\caption{An overview of the proposed \method. In the training phase, the teacher is fixed. The knowledge bridge module (KBM) is trained under the supervision of both the teacher and the student. The student is to mimic the output of the KBM. In the inference phase, we reconstruct the teacher network with low-level layers and KBM. In this way, we obtain a well-trained student network as well as a retrained teacher network.}
	\label{fig:framework}
\end{figure*}

\subsection{Semi-Online Knowledge Distillation}
\label{sec:sokd}
Based on above analysis, the training mechanism that the teacher and the student are updated simultaneously in DML is an effective way for the student to learn similar behaviors of the teacher. Besides, the well-trained teacher in KD provides reliable and stable signals for the student. In this paper, we propose to integrate KD with DML in a unified framework. To this end, we propose a Semi-Online Knowledge Distillation (\method) framework that effectively improves the performance of the student and the teacher. The overview of our framework is shown in figure~\ref{fig:framework}, which consists of three parts, well-trained teacher (low-level and high-level layers), knowledge bridge module (KBM), and student. The KBM is proposed to build a knowledge bridge between the teacher and the student. We take the outputs of low-level layers as the inputs of KBM. More details about KBM will be discussed in section~\ref{sec:kbm}. 

In the training phase, the KBM tries to fully exploit the knowledge of the teacher and transfer it to the student. Meanwhile, the KBM also gains the feedback from the student. Specifically, we do not update the parameters of the teacher since it would destroy the inherent knowledge of the teacher. Instead, we update the parameters of the KBM and the student simultaneously, which is a more tolerant way for the student to approach what the teacher expresses. The training objectives of the KBM and the student can be written as:
\begin{equation}
\begin{split}
    \mL^{kbm}&=\alpha_1\mL_{ce}^{kbm}+\alpha_2KL(p^{kbm},p^t)+ \alpha_3KL(p^{kbm},p^s), \\
    \mL^{s}&=\lambda_1\mL_{ce}^s+\lambda_2KL(p^s,p^{kbm}),
\end{split}
\end{equation}
where $\alpha_1$, $\alpha_2$, $\alpha_3$, $\lambda_1$, $\lambda_2$ are balanced hyperparameters. KL divergence is computed by Eqn.~\ref{eq:kl}. In this sense, we are able to obtain a compact but effective student network. Besides, we also get a more powerful teacher by reconstructing the teacher network with low-level layers and KBM.

\subsection{Knowledge Bridge Module}\label{sec:kbm}
In this subsection, we discuss how to build the KBM, the structure of which is motivated by characteristics of Convolutional Neural Network (CNN). CNN can be roughly composed of low-level and high-level layers. Given a face image, the features detected by low-level layers are usually some fundamental patterns, \eg, edge, corner, and color, which are not sensitive to classes~\cite{albawi2017understanding}. By contrast, the features extracted by high-level layers may be a full face, which is class-specific and related to the output closely. In other words, the discriminative power of a CNN is mainly reflected in the high-level layers~\cite{zeiler2014visualizing}. 

Based on the above theoretical analyses, we design the KBM to be the same as the teacher's high-level layers while using its low-level layer's output as input. Formally, given a pre-trained teacher model $M^t$, our main goal is to obtain a compact yet efficient student model $M^s$. For simplicity, low-level and high-level layers are denoted by $\left\{M_i^t\right\}_{i=1}^l$ and $\left\{M_i^t\right\}_{i=l+1}^L$, respectively. To build a knowledge bridge between $M^t$ and $M^s$, we propose the KBM $M^{kbm}$. The KBM is the same as $\left\{M_i^t\right\}_{i=l+1}^L$ and takes the output of $\left\{M_i^t\right\}_{i=1}^l$ as input.

The main characteristics of the KBM are: 1) Since KBM is the same as high-level layers of the teacher, it has the ability to approach what the teacher expresses. 2) Since different networks have similar representations in low-level layers, sharing the low-level layers would not greatly affect the performance of the KBM. Conversely, it is similar to layer-wise training~\cite{bengio2006greedy, wang2018progressive}, which boosts training stability. Besides, it only brings a small amount of extra calculations.

\begin{table*}[t]
\begin{center}
\resizebox{0.98\textwidth}{!}{
\begin{tabular}{c|cccccccc}
			\toprule 
			Teacher & WRN-28-4&WRN-40-2 & resnet110 & vgg13 & MBNV2 (1.4) & ShuffleNetV1 & resnet32$\times$4 & WRN-28-4  \\
			Student &WRN-16-2& WRN-40-1 & resnet20 & vgg8 & MBNV2 (0.5) & MBNV2 (0.5) & ShuffleNetV1 & resnet56 \\
			\midrule
			Teacher&78.91 & 76.22 & 74.00 & 75.29 & 69.57 & 71.85 & 78.92 & 78.91 \\
			Student &73.32& 71.78 & 69.20 & 70.99 & 65.17 & 65.17 & 71.85 & 73.05 \\
		    \midrule
		    KD~\cite{hinton2015distilling}& 74.72 & 73.54 & 70.73 & 73.58 & 67.94 & 68.19 & 73.90 & 74.21 \\
			AT~\cite{Zagoruyko2017PayingMA}& 74.01 & 72.94 & 70.78 & 72.78 & 66.26 & 65.55 & 73.15 & 73.64 \\
			SP~\cite{tung2019similarity}& 73.81 & 73.18 & 70.69 & 73.58 & 61.52 & 67.67 & 74.87 & 74.28 \\
			CC~\cite{peng2019correlation}& 73.73 & 72.22 & 70.30 & 71.49 & 65.39 & 65.16 & 72.41 & 73.16 \\
			VID~\cite{ahn2019variational} &73.93& 72.52 & 70.26 & 72.13 & 65.15 & 65.49 & 73.10 & 73.24 \\
			CRD~\cite{tian2019contrastive}& 75.05 & 73.57 & 70.88 & 73.24 & 66.47 & 67.32 & 73.76 & 74.34\\
			SSKD~\cite{xu2020knowledge}& 75.57 & 74.02 & 70.70 & 73.96 & 68.37 & 67.99 & 73.99 & 75.31 \\
			SemCKD~\cite{chen2020cross}& 73.63 & 73.32 & 70.41 & 73.35 & 67.70 & 67.19 & 74.29 & 73.13\\
			\midrule
			\method (Teacher) &
		    80.54 &
		    78.41 & 76.32 & 77.16 & 71.83 & 73.41 & 80.46 & 80.47\\
			
		    \method (Student) &
		    \textbf{76.82} &
		    \textbf{75.35} & \textbf{72.08} & \textbf{74.27} & \textbf{69.28} & \textbf{69.50} & \textbf{75.87} & \textbf{75.94}\\
		
		    \bottomrule
		\end{tabular}}
\end{center}
	\caption{Comparison with state-of-the-art offline methods on CIFAR-100 dataset.
	}
		\label{tab:cifar-100}
\end{table*}

\begin{table*}[t]
\begin{center}
\resizebox{0.98\textwidth}{!}{
\begin{tabular}{cc|cc|cc|cc|cc|cc}
			\toprule 
			\multicolumn{2}{c|}{Networks} & \multicolumn{2}{c|}{Vanilla} & \multicolumn{2}{c|}{DML~\cite{Zhang2018DeepML}} & \multicolumn{2}{c|}{KDCL~\cite{Gou2020KnowledgeDA}} & \multicolumn{2}{c|}{DCM~\cite{yao2020knowledge}} & \multicolumn{2}{c}{\method (Ours)}  \\
			\midrule
			Teacher & Student & Teacher & Student & Teacher & Student & Teacher & Student & Teacher & Student & Teacher & Student \\
			\midrule
			WRN-16-2 & WRN-16-2 & 73.32 & 73.32 & 74.46 & 74.56 &74.64 &74.57 & 74.88 & 74.86 & 75.54 & \textbf{75.63}\\
			WRN-28-2 & WRN-28-2 & 75.75 & 75.75 & 76.83 & 77.00 &77.36 &77.32 & 77.71& 77.67 &78.04&\textbf{78.61} \\
			\midrule
			WRN-28-4 & WRN-16-2 & 78.91 & 73.32&79.69&75.27&79.89&75.14&78.92&75.44&\textbf{80.54}&\textbf{76.82}\\
			WRN-28-4 & MobileNet & 78.91 & 73.23 & 80.49&76.94 & 80.64 &76.03&\textbf{80.67}&78.11&80.62&\textbf{78.84}\\
		    \bottomrule
		\end{tabular}}
\end{center}
	\caption{Comparison with state-of-the-art online methods on CIFAR-100 dataset. We refer to the smaller network in online methods as student, otherwise as teacher.
	}
		\label{tab:cifar-100-online}
\end{table*}

\subsection{Extension to Feature-Based Methods}
\label{sec:extension}
In this subsection, we investigate the possibility of extending our method to feature-based methods. The main differences between feature-based and logit-based methods lie in the definition of the knowledge and learning objectives of the student. Feature-based methods learn from the teacher by minimizing the distance (\eg, $L_1$, $L_2$ distance) of intermediate activations. Different methods have different requirements in the positions of intermediate activations. For example, SP only uses the activation of last convolutional layer, while AT adopts activations at the end of each feature extractor block (\eg, residual block). However, it would not be an obstacle of the extension of our method. Our proposed KBM structure can be easily adjusted according to the intermediate activations without loss of performance. The KBM with different blocks will be discussed in detail in section~\ref{exp:bridge}. Formally, the overall training objectives can be modified as follows:
\begin{equation}
\begin{split}
    \mL^{kbm}=&\alpha_1\mL_{ce}^{kbm}+\alpha_2d(T_{kbm}(F_{kbm}),T_{t}(F_{t}))+\alpha_3d(T_{kbm}(F_{kbm}),T_{s}(F_{s})), \\
    \mL^{s}=&\lambda_1\mL_{ce}^s+\lambda_2d(T_s(F_s),T_{kbm}(F_{kbm}))),
\end{split}
\end{equation}
where $\alpha_1$, $\alpha_2$, $\alpha_3$, $\lambda_1$, $\lambda_2$ are balanced hyperparameters. $F$ denotes the feature, $T(\cdot)$ is feature transform function, and $d(\cdot)$ is the distance function.

\section{Experiments}

\subsection{Experimental Details}
\textbf{Choices of Teacher and Student.} We consider six types of networks, including WideResNet~\cite{zagoruyko2016wide}, resnet~\cite{he2016deep}, vgg~\cite{simonyan2014very}, MobileNet~\cite{howard2017mobilenets}, MobileNetV2~\cite{sandler2018mobilenetv2}, and ShuffleNetV1~\cite{zhang2018shufflenet}. For convenience, we use WRN-d-w to denote the WideResNet with depth $d$ and width factor $w$ and MBNV2 (w) to denote MobileNetV2 with a width multiplier of $w$. We use resnet$d$ and ResNet$d$ to represent CIFAR-style and ImageNet-style resnet with depth $d$, respectively. Besides, we use the following rules to construct teacher-student pairs: 1) the teacher and the student have the same architectural style (\eg, vgg13 and vgg8); 2) the teacher and the student are different architectures (\eg, WRN-28-4 and resnet56).

\textbf{Implementation Details.} 
We implement our methods on PyTorch~\cite{paszke2019pytorch}. We run compared methods based on two knowledge distillation benchmarks~\footnote{\href{https://github.com/AberHu/Knowledge-Distillation-Zoo}{https://github.com/AberHu/Knowledge-Distillation-Zoo; }\href{https://github.com/HobbitLong/RepDistiller}{https://github.com/HobbitLong/RepDistiller}} and author-provided codes. 1) For CIFAR-100 dataset, we train all networks for 200 epochs. We use an SGD optimizer with a mini-batch size of 128, the momentum of 0.9, and the weight decay of 5e-4. The learning rate is initialized by 0.1 (MobileNetV2 and ShuffleNetV1 by 0.05) and divided by 10 at both 100 and 150 epochs. 2) For ImageNet dataset, we follow the standard ImageNet parallel training practice on PyTorch. We set the batch size to 256 and train the student model for 100 epochs. The initial learning rate is 0.1 and decayed by 10 at 30, 60 and 90 epochs, respectively. We set $\lambda_1=\lambda_2=\alpha_1=\alpha_2=\alpha_3=1$, $\tau=3$ for our \method on both CIFAR-100 and ImageNet. Code can be found at \href{https://github.com/swlzq/Semi-Online-KD}{https://github.com/swlzq/Semi-Online-KD}.

\begin{table*}[t]
\begin{center}
\resizebox{0.98\textwidth}{!}{
\begin{tabular}{c|cc|ccccccc|cc}
\toprule 
        & Teacher & Student &
        KD~\cite{hinton2015distilling} &
          AT~\cite{Zagoruyko2017PayingMA} & SP~\cite{tung2019similarity} & CC~\cite{peng2019correlation} &
          CRD~\cite{tian2019contrastive} &
          SSKD~\cite{xu2020knowledge} &
          SemCKD~\cite{chen2020cross}$^\dagger$ & 
          \method (Teacher) & \method (Student)\\
			\midrule
			Top-1 & 26.69 & 30.25 & 29.34 & 29.30 & 29.38 & 30.04 & 28.83 & 28.38 & 29.13 & 25.85 & \textbf{28.04} \\
			Top-5 & 8.58 & 10.93 & 10.12 & 10.00 & 10.20 & 10.83 & 9.87 & 9.33 & - & 7.89 &  \textbf{8.92} \\
			\bottomrule
		\end{tabular}}
\end{center}
	\caption{Test error ($\%$) of different offline methods on ImageNet. $^\dagger$ SemCKD~\cite{chen2020cross} only reports Top-1 error.}
		\label{tab:ImageNet}
\end{table*}

\begin{table}[t]
\begin{center}
\resizebox{0.6\textwidth}{!}{
\begin{tabular}{c|c|ccc|c}
			\toprule 
			&Method  & Vanilla & DML~\cite{Zhang2018DeepML} & DCM~\cite{yao2020knowledge} & \method \\
			\midrule
			Teacher&ResNet50 & 23.87 & 24.23 & \textbf{23.43}& 23.49\\
			Student&ResNet18 & 30.25 & 28.98 & 28.65 & \textbf{28.44} \\
		    \bottomrule
		\end{tabular}}
\end{center}
	\caption{Test error (\%) of different online methods on ImageNet. We refer to the smaller network in online methods as student, otherwise as teacher.}
		\label{tab:ImageNet-online}
\end{table}

\subsection{Comparisons with Offline Methods}\label{offline}
\textbf{Results on CIFAR-100.}
In this experiment, we compare our proposed \method with KD~\cite{hinton2015distilling}, AT~\cite{Zagoruyko2017PayingMA}, SP~\cite{tung2019similarity}, CC~\cite{peng2019correlation}, VID~\cite{ahn2019variational}, CRD~\cite{tian2019contrastive}, SSKD~\cite{xu2020knowledge} and SemCKD~\cite{chen2020cross} on eight teacher-student combinations. From Table~\ref{tab:cifar-100}, our \method outperforms all of compared methods by a large margin on all teacher-student pairs, which demonstrates the effectiveness and generalization of our method. It is worth noting that our \method not only achieves high performance in terms of the student's accuracy, but also improves the performance of the teacher. Specifically, the teacher of our SOKD outperforms the vanilla pretrained teacher by 1.87\% Top-1 accuracy, ranging from 1.54\% to 2.32\%. Similar to Vanilla KD method, our \method only leverages logit-based knowledge while showing superior performance than existing state-of-the-art feature-based methods. We show that the knowledge of logit still has huge room for improvement, which is usually underestimated by existing methods. 

\textbf{Results on ImageNet.} In this experiment, we evaluate our proposed \method on ImageNet dataset. Constrained by limited computational resources, we only use ResNet34 and ResNet18 as teacher and student network, respectively. As shown in Table~\ref{tab:ImageNet}, our proposed \method outperforms other state-of-the-art methods. Specifically, our \method outperforms KD of 1.30\% in terms of Top-1 accuracy. Besides, the teacher of \method achieves 0.84\% higher Top-1 accuracy than vanilla teacher. These results show the generalization of our proposed \method in the large-scale dataset.

\subsection{Comparisons with Online Methods}\label{online}
\textbf{Results on CIFAR-100.} In this experiment, we compare our \method with three state-of-the-art online KD methods (\ie, DML~\cite{Zhang2018DeepML}, KDCL~\cite{Gou2020KnowledgeDA}, DCM~\cite{yao2020knowledge}). We consider two training scenarios: 1) training two models with same backbone; 2) training two models with different backbones. We show the results in Table~\ref{tab:cifar-100-online}. From this table, we have the following observations: First, our \method gains the best student models. Second, it is interesting to find that \method further improves the accuracy of the teacher models, which never occurs in traditional offline KD methods. Specifically, our \method evenly achieves 1.96\% higher Top-1 accuracy on four teacher-student pairs. Furthermore, although our main goal is to obtain a small and effective student model, we can further improve teacher by simply adjusting the structure of KBM, which will be discussed in section \ref{further}.

\textbf{Results on ImageNet.} In this experiment, we conduct experiments with two teacher-student pairs. We do not compare with KDCL~\cite{Gou2020KnowledgeDA} due to unavailable code on ImageNet. As shown in Table~\ref{tab:ImageNet-online}, our \method performs better than all compared methods. Specifically, for ResNet50-ResNet18 pair, \method has increased the accuracy of vanilla student and teacher by 1.81\% and 0.38\%, respectively. These results show effectiveness of our \method.

\begin{table}[t]
\begin{center}
\resizebox{0.95\textwidth}{!}{
\begin{tabular}{c|ccccccccc|c}
			\toprule
			Method & Vanilla &
			AT~\cite{Zagoruyko2017PayingMA} &CRD~\cite{tian2019contrastive}&SSKD~\cite{xu2020knowledge}&SemCKD~\cite{chen2020cross}&KDCL~\cite{Gou2020KnowledgeDA}&DCM~\cite{yao2020knowledge}&KD~\cite{hinton2015distilling}&DML~\cite{Zhang2018DeepML}& \method \\ 
			\midrule
			CKA & 0.7200&0.7287&0.7359&0.7773&0.7105&0.8815 &0.8834&0.7966&0.8821&\textbf{0.8890}\\
			IER ($\%$)
			&23.99& 23.81 & 21.31 & 19.87& 24.18 & 20.39 & 16.89& 22.77& 17.15&\textbf{16.32}\\
			MR ($\%$) & \textbf{9.97} & 10.03 & 11.09& 11.59& 10.13& 11.04& 12.91& 10.18& 13.24& 11.61\\
			\bottomrule	
		\end{tabular}}
\end{center}
	\caption{CKA similarity~\cite{kornblith2019similarity}, imitation error rate (IER), and misleading rate (MR) on CIFAR-100 dataset. We take WRN-28-4 and WRN-16-2 as the teacher and the student, respectively.}
		\label{tab:similarity}
\end{table}

\begin{table}[t]
    \centering
    \resizebox{0.98\textwidth}{!}{
    \begin{tabular}{c|c|c|ccc|ccc|ccc}
        \toprule
        &Method&Vanilla&AT&DML-AT&SO-AT&SP&DML-SP&SO-SP&CRD&DML-CRD&SO-CRD\\
        \midrule
        Teacher&vgg13&75.29& 75.29&75.23&\textbf{75.88}& 75.29& 75.49& \textbf{76.68} & 75.29 & 74.62& \textbf{76.41}\\
        Student&vgg8& 70.99& 72.76&71.51&\textbf{73.04}& 73.58& 70.94& \textbf{74.29} & 73.24& 72.62& \textbf{73.74}\\
        \midrule
        Teacher&WRN-28-4& 78.91 & 78.91&78.78&\textbf{80.53}& 78.91& 79.53& \textbf{80.00}& 78.91& 79.03& \textbf{79.42}\\
        Student&resnet56& 73.05& 73.25&73.26&\textbf{73.88}& 74.28& 73.15& \textbf{74.52}& 74.19 & 73.85 & \textbf{74.55}\\
        \bottomrule
    \end{tabular}}
    	\caption{Test accuracy ($\%$) of teacher and student on different training frameworks on CIFAR-100 dataset.}
    	\label{tab:feature-based}
\end{table}

\subsection{Further Experiments}\label{further}

\textbf{Imitation Ability of Student.}
In this experiment, we try to analyze the degree to which the student imitates the teacher.
From Table~\ref{tab:similarity}, we observe that 1) online methods (\ie, DML, KDCL, DCM) generally have learned similar representations to the teacher (higher CKA and lower IER) while suffering more serious misleading (higher MR). 2) our \method achieves much higher CKA and lower IER than all of compared methods, revealing that our \method has stronger ability to capture the behaviors of the teacher. Meanwhile, the supervision signals provided by the teacher in \method are much more accurate (lower MR) than that in DML. The main reason is that the KBM is able to obtain valuable information from the well-trained teacher. 

\textbf{Extensions to Feature-Based Methods.}
To investigate the scalability of our \method, we extend it to feature-based methods. Since feature information is related to network structure, we choose two different combinations of the teacher and the student in this experiment. Meanwhile, we also try to apply the training framework of DML into different methods. For simplicity, we use ``DML/SO-M” to denote training M method with DML/SOKD framework. From Table~\ref{tab:feature-based}, the performance of feature-based methods significantly decreases when applied with DML framework on most cases but our \method performs better than baselines. One possible reason is that in the early training of DML the representations might be meaningless, which would mislead the optimization directions of the teacher and the student. Instead, the training of \method is supervised by a well-trained teacher, which has learned powerful representations.

\begin{table}[t]
    \centering
    \resizebox{0.98\textwidth}{!}{
    \begin{tabular}{c|c|cc|ccccc}
        \toprule
         &Method & vanilla & KD~\cite{hinton2015distilling} & classifier&classifier + 1 block&classifier + 2 blocks&classifier + 3 blocks&whole network \\ 
         \midrule
         Teacher&WRN-28-4 & 78.91 &78.91& 79.35& 80.47& 80.64&\textbf{81.08}& 80.96\\
         Student&resnet56 & 73.05& 74.21& 74.88& \textbf{75.94}& 75.90& 75.87& 75.34\\
         \bottomrule
    \end{tabular}}
    \caption{Effect of different KBMs on CIFAR-100 dataset. The last row ``whole network" means the KBM is the same as the whole teacher network (It refers to WRN-28-4 here).}
		\label{tab:num-blocks}
\end{table}

\begin{table}[t]
    \centering
    \resizebox{0.6\textwidth}{!}{
     \begin{tabular}{ccc|cc}
            \toprule
            $\mL_{ce}^{kbm}$ & $KL(p^{kbm},p^t)$ & $KL(p^{kbm},p^s)$ & WRN-28-4 (Teacher) & resnet56 (Student) \\
            \midrule
            $\surd$ &  &  & 78.02 & 75.45 \\
            $\surd$ & $\surd$ &  & 79.24 & 74.91 \\
            $\surd$ &  & $\surd$ & 79.34 & 74.87 \\
            & $\surd$ & $\surd$  & 80.03 & 75.56 \\
            $\surd$& $\surd$ & $\surd$  &  \textbf{80.47} & \textbf{75.94} \\
			\bottomrule
		\end{tabular}}
   	\caption{Effect of different losses for KBM on CIFAR-100 dataset.}
		\label{tab:ablation-loss}
\end{table}

\textbf{Effect of Different KBMs.}
\label{exp:bridge}
In this experiment, we investigate the effect of different KBMs. 
We design the KBM with last classifier layers and last n feature extractor blocks of the teacher.
As shown in Table~\ref{tab:num-blocks}, all KBMs achieve better performance than KD, which demonstrates the effectiveness and flexibility of our \method. The KBM can be changed according to different demands while maintaining the performance. However, the KBM with only classifier performs much worse than those with several extractor blocks, which means that it is necessary to adjust the parameters of intermediate layers such that the student is easier to learn similar representations from the teacher. However, taking the whole WRN-28-4 as KBM does not lead to further improvement, which shows the necessity of sharing teacher's low-level layers. For the purpose of reducing computational cost, in this paper, we set the structure of KBM to be the same as the last feature extractor block and last classifier in most cases.

\textbf{Effect of Different Losses for KBM.}
We investigate the effect of different losses for KBM. The results are shown in table~\ref{tab:ablation-loss}. When only equipped with $\mL_{ce}^{kbm}$, the performance of the student is improved. The main reason is that KBM and student are trained simultaneously, the student is easier to imitate the KBM than the well-trained teacher. However, only adding the supervision of the teacher (\ie, w/o $KL(p^{kbm},p^s)$) or only obtaining the feedback from the student (\ie, w/o $KL(p^{kbm},p^t)$) both lead to performance degradation, which shows the indispensability of reliable signals and real-time feedback. Equipped with all losses, both teacher and student achieve highest accuracy, which shows the effectiveness of proposed losses for the KBM.

\section{Conclusion}
In this paper, we have discovered that KD and DML are complementary to each other on knowledge distillation task. Inspired by this observation, we have proposed a Semi-Online Knowledge Distillation (\method) method that integrates KD with DML. Our \method have effectively leveraged the well-trained teacher's supervision signals and alleviated the imitation difficulty of the student. Moreover, we have extended the proposed framework to feature-based distillation methods, which also achieved competitive performance. Extensive experiments on two benchmarks have shown that our method consistently outperforms state-of-the-art methods on knowledge distillation tasks with different teacher-student pairs.

\section*{Acknowledgement}
This work was supported by the Key Realm R\&D Program of Guangzhou (202007030007), China Scholarship Council (CSC).

\bibliography{egbib}
\end{document}